\documentclass{article}
\usepackage{ijcai23}

\usepackage{times}
\usepackage{soul}
\usepackage{url}
\usepackage[utf8]{inputenc}
\usepackage[small]{caption}
\usepackage{graphicx}
\usepackage{amsmath}
\usepackage{amsthm}
\usepackage{booktabs}
\usepackage{algorithm}
\usepackage{algorithmic}
\usepackage[switch]{lineno}
\usepackage{indentfirst}
\usepackage{arydshln}
\usepackage{amssymb}
\usepackage{pifont}
\usepackage{listings}

\newcommand{\cmark}{\ding{51}}%
\newcommand{\xmark}{\ding{55}}%

\title{Beyond Classification: Financial Reasoning in State-of-the-Art Language Models}

\author{
Guijin Son$^{1,2,6}$
\and
Hanearl Jung$^2$\and
Moonjeong Hahm$^{3}$\and
Keonju Na$^4$\And
Sol Jin$^5$
\affiliations
$^1$Yonsei University $^2$OneLineAI $^3$Chung-Ang University\\
$^4$Seoul National University of Science and Technology
$^5$Seoul National University
$^6$MODULABS
\emails
spthsrbwls123@yonsei.ac.kr,
earl@onelineai.com,
daily6298@cau.ac.kr,
keonju2@seoultech.ac.kr,
jinsol9770@snu.ac.kr
}

\begin{document}

\maketitle

\begin{abstract}
Large Language Models (LLMs), consisting of 100 billion or more parameters, have demonstrated remarkable ability in complex multi-step reasoning tasks. However, the application of such generic advancements has been limited to a few fields, such as clinical or legal, with the field of financial reasoning remaining largely unexplored. To the best of our knowledge, the ability of LLMs to solve financial reasoning problems has never been dealt with, and whether it can be performed at any scale remains unknown.  To address this knowledge gap, this research presents a comprehensive investigation into the potential application of LLMs in the financial domain. The investigation includes a detailed exploration of a range of subjects, including task formulation, synthetic data generation, prompting methods, and evaluation capability. Furthermore, the study benchmarks various GPT variants with parameter scales ranging from 2.8B to 13B, with and without instruction tuning, on diverse dataset sizes. By analyzing the results, we reveal that the ability to generate coherent financial reasoning first emerges at 6B parameters, and continues to improve with better instruction-tuning or larger datasets. Additionally, the study provides a publicly accessible dataset named sFIOG (Synthetic-Financial Investment Opinion Generation), consisting of 11,802 synthetic investment thesis samples, to support further research in the field of financial reasoning. Overall, this research seeks to contribute to the understanding of the efficacy of language models in the field of finance, with a particular emphasis on their ability to engage in sophisticated reasoning and analysis within the context of investment decision-making.  We release our models, dataset, and code \footnote{https://github.com/guijinSON/FIOG/tree/main}.
\end{abstract}

\section{Introduction}
Large Language Models(100+ billion parameters) have undergone remarkable advancements in recent years, enabling them with the ability to generate coherent and meaningful text~\cite{wei2022emergent}.  These LLMs have demonstrated notable abilities in performing complex multi-step reasoning, either by thinking "step by step"~\cite{kojimalarge} or leveraging Chain-of-Thought(CoT) prompts~\cite{wei2022chain}.  Various fields have attempted to harness such reasoning ability, and among them, the field of clinical research has made notable progress by developing domain-specific LLMs like Med-Palm~\cite{singhal2022large}, retrained on massive amounts of domain-specific texts and tasks, which achieves performance comparable to that of human clinicians.  In situations where data is insufficient to train dedicated language models, researchers have directed their efforts towards developing advanced prompt engineering techniques, such as Legal Prompt Engineering (LPE)~\cite{trautmann2022legal}, or generation of synthetic data via LLMs and training of smaller language models on such samples~\cite{yunxiang2023chatdoctor}.  However, there is a lack of comprehensive investigation for either of the methods in the financial domain, leaving the field of financial reasoning largely unexplored.

The research of natural language processing in the financial domain has predominantly been confined to token or sequence classification tasks~\cite{araci2019finbert,shah2022flue}. This is likely due to the lack of datasets or tasks suitable for training generative language models. Even dedicated financial language models like BloombergGPT, tend to prioritize tasks such as sentiment analysis, binary classification, and named entity recognition, with limited attention given to numerical reasoning tasks~\cite{wu2023bloomberggpt}.

Our research aims to comprehensively investigate the \textbf{financial reasoning} capabilities of language models, specifically their ability to generate logically coherent and persuasive investment opinions. The investigation involves both prompt engineering and specialized training of smaller language models~\cite{fu2023specializing}, seeking to advance our understanding on the ability of language models to engage in sophisticated reasoning and analysis within the context of investment decision-making. Accordingly, our research introduces an original financial reasoning task called "Financial Investment Opinion Generation (FIOG)", which involves the generation of investment opinions by language models with either parametric or injected knowledge. We then benchmark various GPT variants, ranging in size from 2.7B to 13B, with and without instruction-tuning~\cite{ouyang2022training}, on the dataset. Additionally, we propose a novel prompting method called In-Context Question Answering for controlled generation of context. Finally, we investigate the alignment between LLM-based evaluators, such as G-Eval~\cite{liu2023gpteval}, and human evaluators for financial texts, in order to gain insights into the efficacy of such evaluators in the financial domain.

To support further research on financial reasoning, we provide a publicly accessible dataset named sFIOG (Synthetic-Financial Investment Opinion Generation), which includes 11,802 synthetic investment opinion samples. This dataset is intended to enable benchmarking and experimentation in the field of financial language modeling and investment opinion generation.

\section{Related Work}
\subsection{Reasoning with Language Models}
Language Models (LMs) trained using conventional pre-training objectives have demonstrated the ability to acquire complex reasoning capabilities once they reach a certain scale~\cite{wei2022emergent}. However, recent research has shown that the parameter requirements for complex reasoning abilities of LMs can be significantly alleviated through a process called instruction tuning~\cite{ouyang2022training}. Further research has suggested that narrowing down the model's focus to specialize in a specific field can result in additional alleviation of parameter requirements. This can be achieved by including task-specific Chain-of-Thought (CoT) data in the instruction-tuning process, allowing the model to acquire specialized reasoning capabilities~\cite{fu2023specializing}. Some researchers have adopted this approach, leveraging domain-specific CoT data, which is often generated by the LLMs themselves, to enable domain-specific reasoning abilities~\cite{yunxiang2023chatdoctor}. However, the effectiveness of this approach across different domains and the potential variability in parameter and data requirements for specific domains remain relatively unexplored. Accordingly, it is plausible that domains characterized by complex nomenclature and reasoning steps, which significantly deviate from general, widely applicable patterns, may necessitate higher parameter and data requirements.

\subsection{Financial Natural Language Processing}

The financial domain has been quick to adopt advancements in generic natural language processing research. Notably, BloombergGPT, a language model with 50 billion parameters specifically dedicated for finance, stands out as a significant development in this field~\cite{wu2023bloomberggpt}.However, despite its significance, BloombergGPT and recent research of the field have limitations in terms of their investigation in reasoning abilities, which have been left out of the scope of research. The focus of predominant research in the financial domain has largely been limited to token or sequence classification tasks~\cite{araci2019finbert,shah2022flue}, likely due to the scarcity of suitable datasets or tasks for training generative language models. For instance, corpora containing financial reasoning steps, which are essential for training language models for tasks such as investment opinion generation, are mostly confidential in nature and therefore excluded from the training data of publicly available language models~\cite{scao2022bloom,black2022gpt,touvron2023llama}. This limitation poses challenges for developing language models with specialized reasoning capabilities in the financial domain. 

Though this study does not involve the development of a finance-native LM of its own, it distinguishes itself from previous research as it comprehensively investigates the circumstances under which specialized financial reasoning capabilities can be enabled.

\section{Task Formulation}

In this paper we introduce a novel task called Financial Investment Opinion Generation(FIOG), the term encompasses all tasks aiming to train or prompt language models to generate investment opinions in the context of finance, leveraging either parametric or injected knowledge. Our variant of the FIOG task involves providing language models with the necessary information as part of the input. The input information in our variant is provided in two types: full-text and question-and-answer (Q\&A). In the full-text type, the input consists of complete text passages, while in the Q\&A type, the input comprises pairs of questions and corresponding answers. The Q\&A type is used to train and prompt our model via In-Context Question Answering, which will be explained later in the paper. Incorporating investment decision-relevant information as part of the input, enables us to investigate the ability of Language Models (LMs) as reasoning engines, rather than knowledge databases, and allows for a more targeted and effective training process.

\section{Dataset Creation}

\begin{table*}[t]
\resizebox{\textwidth}{!}{%
\begin{tabular}{ll|ll|lll}
\hline
\multicolumn{2}{l|}{\textbf{Investment Opinion}} & \multicolumn{2}{l|}{\textbf{(RE) Full-Text Type}} & \multicolumn{3}{l}{\textbf{(RE) Q\&A Type}}        \\ \hline
Coverage       & Investment Thesis      & Full-Text     & Investment Thesis      & Question & Q\&A Pair & Investment Thesis \\ \hline
752            & 1,087                   & 1,087         & 4,386                  & 10,437   & 26,138    & 11,802              \\ \hline
\end{tabular}%
}
\caption{Dataset Overview. (RE) denotes that the set has been regenerated.}
\label{tab:2}
\end{table*}

To support further research on financial reasoning, we provide a publicly accessible dataset named sFIOG (Synthetic-Financial Investment Opinion Generation). The sFIOG dataset is generated through the following steps.
\begin{enumerate}
  \item Collection of expert-written analyst reports: We gathered 1,087 analyst reports from various sources, including J.P Morgan, Truist Financial Corp, and Oppenheimer \& Co. These reports cover 752 companies in the U.S stock market.
  \item Expert-Written investment thesis set construction: We extracted the "Investment Thesis" and "Related Risk" sections from each analyst report, resulting in a set of expert-written investment theses.
  \item Full-Text type input construction: We constructed the Full-Text type input by collecting the abstract from each analyst report.
  \item Q\&A type input question generation: Using the GPT3.5-Turbo API, we fed the Full-Text type input and required it to generate questions addressing important information.
  \item Dummy answer generation: We used the GPT3.5-Turbo API to generate dummy answers for the questions generated in step 4. Human annotators were hired to eliminate answers that deviated greatly from reality.
  \item Investment opinion generation: The GPT3.5-Turbo API was employed to generate investment opinions for both types of inputs.
\end{enumerate}

In step 4, we extract questions from a given text rather than relying solely on a LLM to few-shot generate questions on a given topic. This approach is expected to generate questions that inquire about information deemed important by human experts rather than generating random questions. For comparison, we also construct a set of few-shot generated questions. To assess the lexical and syntactic diversity of each method, we use three metrics: Mass and HD-D for lexical diversity, and Syntactic Sim. for syntactic diversity. Mass and HD-D are established metrics for measuring lexical richness and have been shown to be reliable across texts of different lengths \cite{torruella2013lexical,mccarthy2010mtld}. A higher HD-D score indicates greater lexical richness, while a higher Mass score indicates the opposite. For syntactic diversity, we use Syntactic Sim., which measures the average pairwise similarity of the dependency tree across generated samples \cite{oya2020syntactic}. A higher Syntactic Sim. value indicates greater similarity in syntactic structures across generated samples. As presented in Table~\ref{tab:1}, our approach resembling question extraction yields synthetic data with a higher degree of both lexical and syntactic diversity.

\begin{table}[ht]
\centering
{%
\begin{tabular}{rrr}
\hline
\textbf{Generation}  & \textbf{Few-Shot} & \textbf{Step 4.} \\ \hline
HD-D                 & 0.811             & 0.873                     \\
Mass                 & 0.034             & 0.025                     \\
Syntactic Sim. & 0.578             & 0.42                      \\ \hline
\end{tabular}%
}
\caption{Quantitative assessment of questions generated via few-shot generation against ours (step 4).}
\label{tab:1}
\end{table}

Step 5, adds multiple dummy answers for the questions generated in the prior step. These dummy answers were carefully screened by a human annotator to eliminate those that deviate greatly from reality. We expect this process to add to the diversity of the dataset aiding the fine-tuning of complex reasoning, similar to diverse reasoning~\cite{ho2022large}.

Table~\ref{tab:2} includes the statistics for the constructed sFIOG dataset. Our dataset encompasses three types of the investment thesis. First, we have 1,087 expert-written investment theses. Second, we have 4,386 investment theses generated with full-text type input. It is noteworthy that the investment thesis generated with the full-text type input exhibits a balanced distribution of buy, hold, and sell opinions, with 1,462 samples for each. Finally, we have 11,802 samples generated with Q\&A type input. Each sample was generated with 13 or more Q\&A pairs, ensuring that a sufficient amount and diversity of information was provided for the language models to formulate comprehensive investment opinions. More than one sample was generated for each set of Q\&A pairs to add to the diversity of the dataset.

The publicly accessible sFIOG dataset is limited to the Q\&A type input subset of the dataset due to the restriction of third-party sharing of the expert-written analyst reports collected from the web. To the best of our knowledge, the publicly accessible version of the sFIOG dataset is comprised only of synthetically generated questions, answers, and investment opinions.

\section{In-Context Question Answering}

Both LLM or their smaller variants have been pointed out to hallucinate, or generate context unfaithful from real world information~\cite{ji2023survey}. Even if these LMs manage to accurately retrieve real-world information that they have memorized during the pre-training stage, there are still risks of the information being outdated or non-stationary~\cite{son2023removing}. To address this issue, we propose In-Context Question Answering, where a list of question-and-answer pairs are provided instead of full-text contexts. Through experiments, we demonstrate that our approach has several advantages compared to previous full-text in-context learning approaches when zero-shot prompting LLMs. A sample of the questions used is presented in \ref{sec:appendix:prompt}.

First, our findings indicate that generations grounded on Q\&A pairs exhibit a higher degree of controlled behavior, or a lower likelihood to generate unintended context, compared to conventional in-context learning generations. For instance, approximately 11.12\% of the samples generated with conventional in-context learning included analysis on the pandemic, even though the investment opinion was intended for the post-pandemic era. In contrast, when using in-context question answering, the chances of generated samples to discuss pandemic-related issues, despite their absence in the provided Q\&A sets, was merely 1.63\%. This suggests that the proposed in-context question answering may be a more effective approach to zero-shot prompt LLMs to generate controlled outputs, making it more suitable for specific contexts and scenarios, such as post-pandemic era financial analysis. We speculate that such behavior is because in-context question answering delivers a refined version of information with most of the irrelevant text removed, resulting in a more concise and focused input. Language models are susceptible to distraction from irrelevant text~\cite{shi2023large}, and the provision of context in a Q\&A format allows them to concentrate on the core information without being influenced by unnecessary or irrelevant sentences. This conciseness and absence of irrelevant text in the Q\&A format may enable language models to better align with the intended task, leading to improved performance and controlled behavior in generating contextually relevant and accurate content.

Second, we conducted a survey with hired human annotators using a subset of 1,000 samples from each type. In order to assess the performance of our LLM-based evaluators in comparison to human annotators, we also conducted the identical survey using GPT-4 as a respondent, following previous research on G-Eval~\cite{liu2023gpteval}. The survey presented respondents with three samples at a time, one from each of the expert-written, full-text type, and Q\&A type. They were then required to answer two questions:

\begin{enumerate}
  \item \emph{Which investment thesis contains the most investment helpful information?}
  \item \emph{Which investment thesis presents a more logically structured and reasonable argumentation?}
\end{enumerate}

\begin{figure}[ht]
    \centering
    \includegraphics[width=\columnwidth]{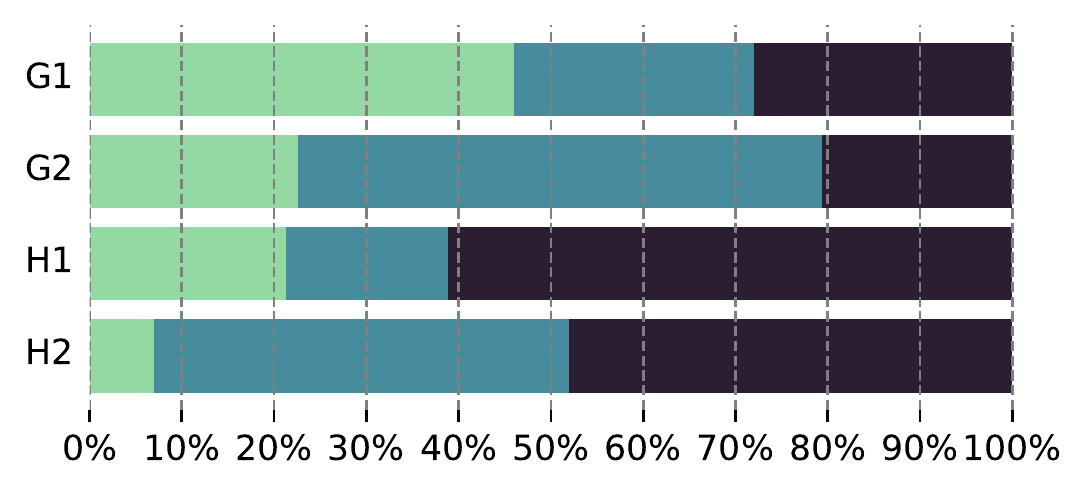}
    \caption{Qualitative Evaluation of Collected Investment Theses: Green denotes expert-written, blue represents full-text type, and dark blue indicates Q\&A type. G1 and G2 refer to GPT-4 answers for Question1 and Question2, respectively. H1 and H2 denote human answers for Question1 and Question2, respectively.}
    \label{survey}
\end{figure}

Figure~\ref{survey}, indicates that human evaluators perceived Q\&A type generation to contain the most investment-helpful information in 61.2\% of cases and demonstrated the most coherent argumentation in 48\% of cases. In contrast, Full-Text type generation was found to have relatively fewer investment-helpful information, which may be attributed to the presence of irrelevant text that could disrupt the language model's output. Notably, the generated samples in either full-text or Q\&A type were preferred by human annotators over the expert-written samples for both questions. We speculate that this preference for generated samples over expert-written thesis may be due to the fact that expert-written thesis are tailored for professionals with domain-specific expertise, and may omit explanations or assumed background knowledge, potentially affecting their comprehensibility to human evaluators. An investigation of the inter-annotator agreement was conducted on a subset of 350 samples for each question, revealing a decent Krippendorff's alpha of 0.63 for question 1 and 0.68 for question 2.

Furthermore, we conduct the identical survey using GPT-4, following G-Eval, we use the following prompt:
\begin{quote}
\textit{You are a professional financial researcher. You will be given an investment thesis. Your task is to rate the thesis on the following metric. Please make sure you read and understand these instructions carefully. Please keep this document open while reviewing, and refer to it as needed.}

\textit{Evaluation Criteria:}

\textit{Investment-Helpfulness (1-5) - the quality and diversity of financial facts provided in the passage. The investment thesis should provide a diverse set of quantitative information. Quantitative information must include numerical values. Concentrate on the diversity and amount of facts provided. Ignore the argumentation for the moment.}

\textit{Financial Argumentation (1-5) - the quality of the financial reasoning and supporting evidence in the passage. This includes the logical coherence of the financial argument, the strength of the financial evidence provided, and the overall persuasiveness of the financial argument. Specifically, this criterion evaluates the effectiveness of the financial analysis and the quality of the financial data used to support the investment thesis.}
\end{quote}

\begin{figure}[ht]
\centering{
\begin{minipage}[t]{0.47\linewidth}
    \centering
    \includegraphics[width=1\textwidth]{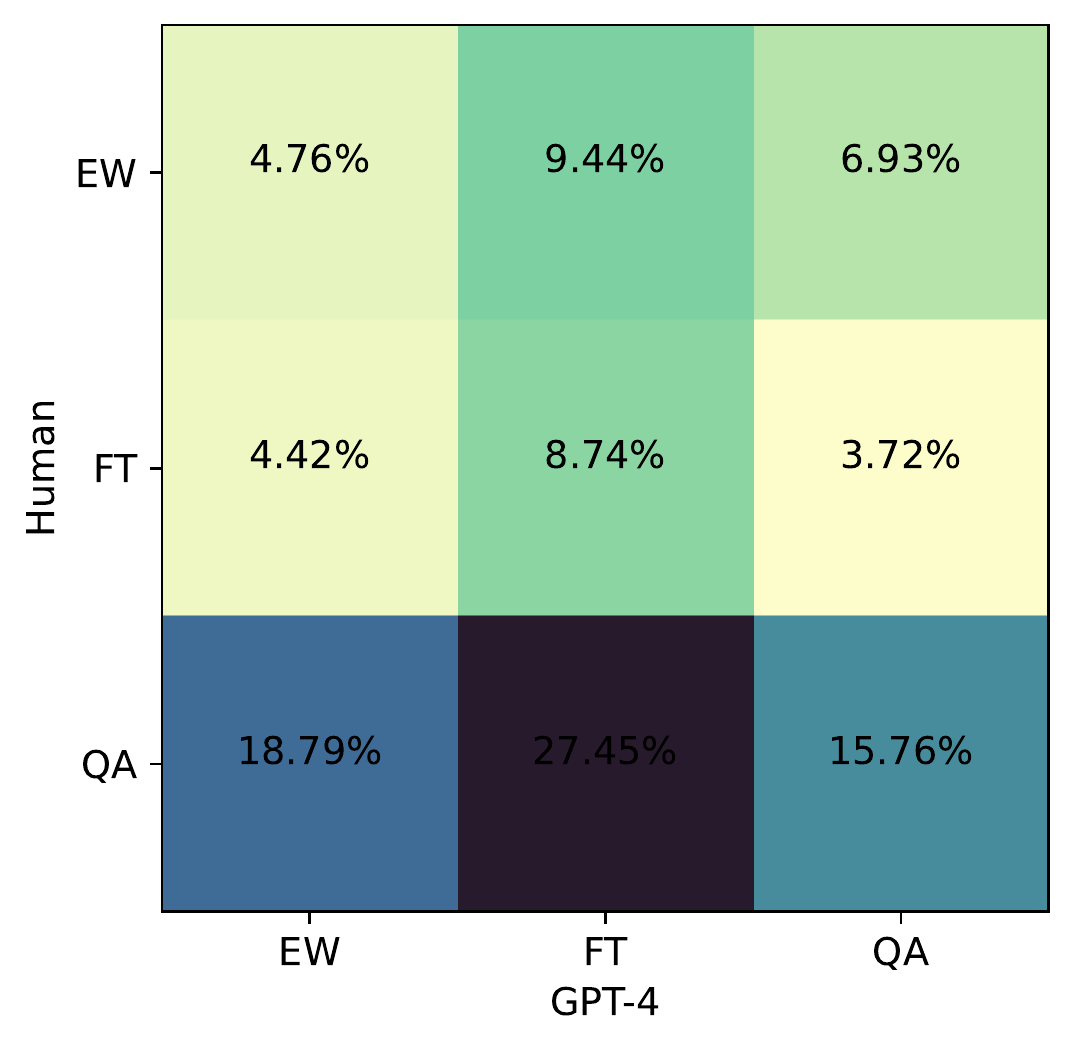}
\end{minipage}
\hspace{0.1cm}
\begin{minipage}[t]{0.47\linewidth} 
    \centering
    \includegraphics[width=1\textwidth]{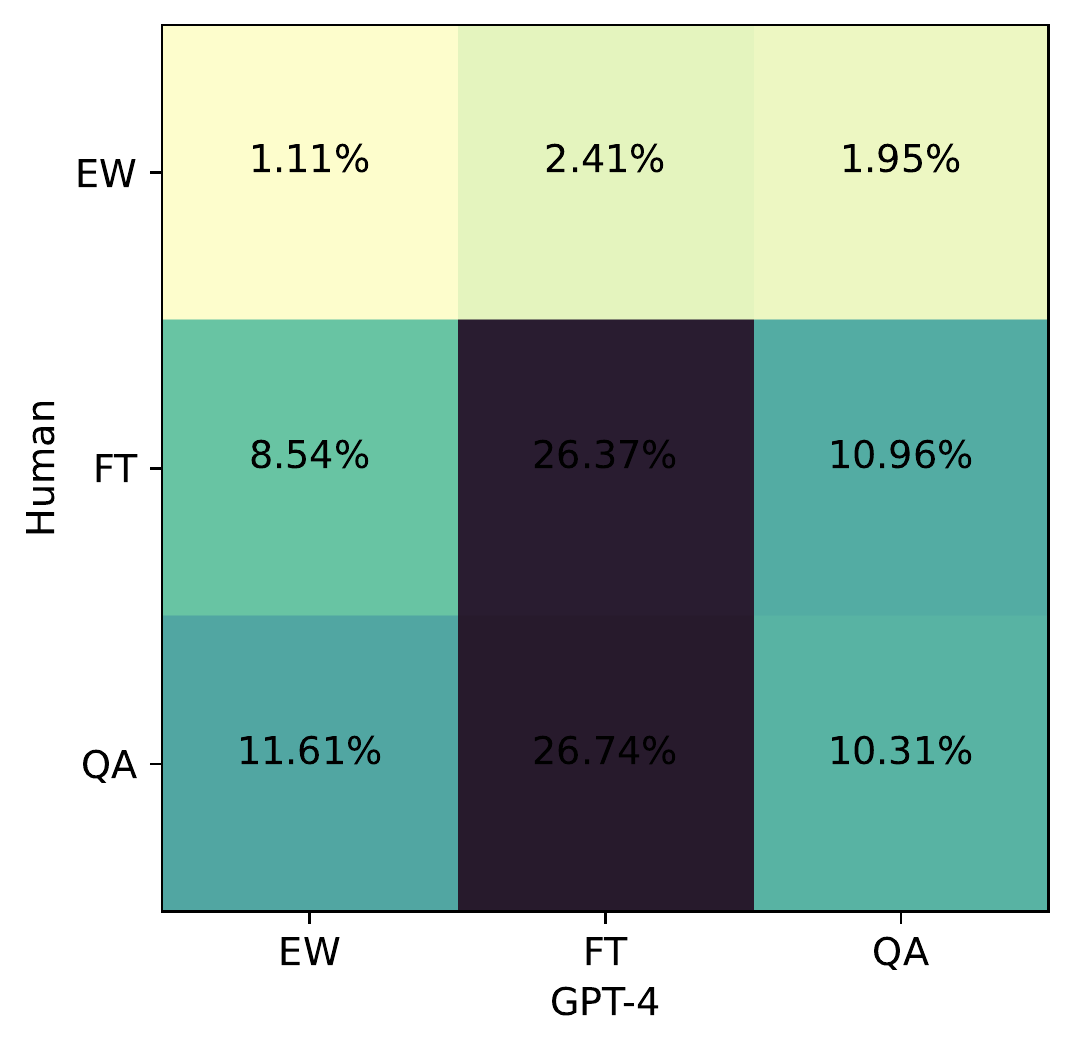}
\end{minipage} 
\caption{Left for Q1, Right for Q2.}
\label{confusion}
}
\end{figure}

The responses from LLMs were compared with the decision of human annotators to investigate the efficacy of LLM applications for the evaluation of financial reasoning. Unlike previous research~\cite{gilardi2023chatgpt}, our study found a notable disparity between GPT-4 and human judgments, with low correlation observed regardless of the presence of CoT explanations. Figure~\ref{confusion} displays the confusion matrix comparing the decisions of human and LLM evaluators. The results indicate that the agreement rate between the two evaluators was only 29.26\%, and 34.6\% for each question correspondingly. Moreover, the Spearman correlation coefficients between human and LLM decisions were -0.07 for question one and -0.073 for question two, significantly lower than that of previous research that reported 0.514~\cite{liu2023gpteval}. This disparity may be attributed to two key factors. First, unlike prior research that focused on LLMs' evaluation of summarization quality or zero-shot classification of tweets, our study required the LLMs to evaluate financial reasoning, which is a more intricate and complex task. Additionally, LLMs were never trained for such tasks, which may have impacted their performance in evaluating the quality of financial reasoning. Secondly, the financial domain poses unique complexities, including diverse nomenclature and domain-specific knowledge, which may present a challenge for generic LLMs to fully comprehend and accurately evaluate the coherence of financial reasoning. Following our findings, LLMs are no longer used as evaluators in this paper.

\begin{table*}[ht]
\resizebox{\textwidth}{!}{%
\begin{tabular}{lcllllllll}
\hline
Base Model &
  \multicolumn{1}{l}{Instruction-Tuning} &
  \multicolumn{4}{l}{ROUGE-L} &
  \multicolumn{4}{l}{BERTScore} \\
 &
  \multicolumn{1}{l}{} &
  type\#1 &
  type\#2 &
  type\#3 &
  average &
  type\#1 &
  type\#2 &
  type\#3 &
  average \\ \hline
LLama &
  \cmark &
  \textbf{0.283} &
  \textbf{0.178} &
  \textbf{0.359} &
  \textbf{0.273} &
  0.830 &
  \textbf{0.849} &
  \textbf{0.855} &
  \textbf{0.845} \\
Galactica &
  \cmark &
  0.108 &
  0.028 &
  0.114 &
  0.083 &
  0.794 &
  0.807 &
  0.799 &
  0.800 \\
GPT-J      & \cmark & 0.159 & 0.023 & 0.183 & 0.122 & \textbf{0.836} & 0.692 & 0.836 & 0.788 \\
Pythia(2.8B) & \cmark & 0.022 & 0.000 & 0.023 & 0.015 & 0.731          & 0.769 & 0.735 & 0.745 \\ \hline
LLama      & \xmark & 0.080 & 0.123 & 0.180 & 0.128 & 0.592          & 0.778 & 0.723 & 0.698 \\
Galactica  & \xmark & 0.086 & 0.027 & 0.097 & 0.070 & 0.777          & 0.804 & 0.773 & 0.785 \\
GPT-J      & \xmark & 0.054 & 0.023 & 0.139 & 0.072 & 0.773          & 0.692 & 0.818 & 0.761 \\
Pythia(2.8B) & \xmark & 0.017 & 0.012 & 0.018 & 0.016 & 0.729          & 0.795 & 0.728 & 0.751 \\ \hline
\end{tabular}%
}
\caption{Results for LLama, Galactica, GPT-J, and Pythia (2.8B), both with and without instruction-tuning, obtained on the sFIOG test dataset. The evaluation was carried out across three distinct subsets. Type\#1 consisted of companies and questions from the training set with new corresponding answers. Type\#2 featured companies from the training set paired with new, previously unencountered question-and-answer combinations. Lastly, Type\#3 introduced companies not present in the training set, accompanied by new question-and-answer pairs.}
\label{main}
\end{table*}

Overall, the aforementioned experiments yield two important findings. Firstly, the results discover that LLMs are inadequate as evaluators for financial reasoning tasks, given the limited alignment observed between LLMs and human evaluators. Secondly, the proposed In-Context Question Answering method represents a promising alternative to traditional prompting methods, exhibiting improved controlledness and generating better-quality reports. Notably, this method could be applicable to a broader range of fields beyond finance, wherever controlled generation of information-rich texts is required.

\section{Experiments}
\subsection{Experimental Setup}

In this research, we assessed four GPT variants (2.8B to 13B parameters) with and without instruction tuning, as detailed in Table~\ref{models}. This comparison aimed to identify the point at which the ability to generate financial reasoning emerges. An example of the generation is presented in \ref{sec:appendix:gen}.

\begin{table}[ht]
\centering{%
\begin{tabular}{lll}
\hline
Base Model & Instruction-Tuned     & Param. \\ \hline
Pythia     & dolly-v2-3b           & 2.8B   \\
GPT-J      & dolly-v1-6b           & 6B     \\
Galactica  & galpaca-6.7b          & 6.7B   \\
LLama      & vicuna-13b-delta-v1.1 & 13B    \\ \hline
\end{tabular}%
}
\caption{Summary of GPT variants employed in the experiments, detailing their parameter sizes and whether they underwent instruction tuning. Checkpoints for instruction-tuned models were imported from HuggingFace.}
\label{models}
\end{table}

Models in this study were trained using Lora~\cite{hu2021lora} and quantization for enhanced hardware efficiency, with a maximum token length of 2048 and an AdamW optimizer. Each model was trained in three epochs on the full sFIOG dataset, which is consisted of 11,802 samples. During the test phase, decoding settings were configured to enhance the quality and diversity of generated outputs, while ensuring a fair comparison across models. The parameters were set as follows: top\_k=50, top\_p=0.95, no\_repeat\_ngram\_size=3, and max\_new\_tokens=512. By setting a fixed maximum number of tokens, we prevented models that generate longer sequences from appearing to outperform others in the evaluation. 

The test dataset for this study is comprised of three distinct subsets to evaluate the performance of the GPT variants in different settings. The first subset included companies and questions that appeared in the training set but with new corresponding answers. The second subset featured companies from the training set but paired with new, previously unencountered question-and-answer combinations. Lastly, the third subset introduced companies that did not appear in the training set, accompanied by new question-and-answer pairs. Through this dataset split we assess the models' capabilities in generating financial reasoning across varying degrees of familiarity and novelty.

To evaluate the generated context, we used both automated metrics and human evaluations. Automated metrics included rouge-2 and rougeL~\cite{lin-2004-rouge}, measuring text overlap, and BERTScore~\cite{zhang2019bertscore}, assessing semantic similarity. As mentioned previously, we excluded LLM-based evaluators due to their misalignment with human judgments.

\subsection{Model Scale and Financial Reasoning}

In Table~\ref{main}, we present the results for LLama~\cite{touvron2023llama}, Galactica~\cite{taylor2022galactica}, GPT-J, and Pythia (2.8B)~\cite{biderman2023pythia}, with and without instruction-tuning, on the sFIOG test dataset. Our findings indicate that the ability to generate coherent investment opinions emerges in models with sizes between 2.8B $\sim$ 6B and continues to improve as the model scales. For instance, LLama demonstrates superior performance, achieving the highest average scores in ROUGE-L (0.217) and BERTScore (0.821). There are two possible explanations for the scaling behavior of financial reasoning abilities in these models: (1) larger models are typically trained on more tokens, thereby accumulating a greater amount of knowledge essential for generating well-informed investment theses, and (2) the architecture of larger models inherently allows for improved reasoning capabilities, enabling them to better analyze and synthesize the information they have learned. Consequently, as model size expands, it leads to a stronger ability to effectively generate financial reasoning, as demonstrated by the superior performance of the LLama model in our experiments. An exception in the scaling behavior is observed between GPT-J and Galactica, with GPT-J surpassing Galactica in performance, despite its smaller size. We posit that this discrepancy may arise from two factors: (1) GPT-J is trained on a substantially larger corpus of tokens (402 billion) from a general domain, while Galactica has been trained on a smaller, science-specific corpus (106 billion); (2) The size difference between the two models is relatively minimal, at just 0.7B. This observation is consistent with recent research, suggesting that training smaller models with an increased number of tokens beyond the chinchilla optimal point can yield improved performance~\cite{touvron2023llama}. Furthermore, this finding emphasizes the potential trade-offs of domain-specific training, which could compromise a model's robustness across broader contexts.

\subsection{Instruction-Tuning and Financial Reasoning}
We observe that instruction-tuning plays a significant role in enhancing the performance of all models across both evaluation metrics. However, the degree of improvement varies among models, which may be due to the difference of instruction-tuning datasets used to fine-tune each model. It is noteworthy that Pythia (2.8B), the smallest model employed in our experiments, failed to demonstrate the ability to generate coherent financial reasoning, even when instruction-tuning was applied. This finding implies that the ability to generate financial reasoning could be an emergent property that becomes evident as the model size exceeds a specific threshold.

\subsection{Dataset and Financial Reasoning}

In examining the performance of the models across each subset of the dataset, we find that the models exhibit their weakest performance in type\#2 questions, which involve companies included in the training set but are accompanied by new question-and-answer pairs. This observation departs from the authors' initial assumption that type\#3 questions, featuring companies not present in the training set, would pose the greatest challenge. The results demonstrate that generating financial opinions for novel question-answer pairs concerning familiar companies is a more demanding task for the models. This finding aligns with past research, suggesting that the non-stationary knowledge acquired during the training process may hinder the models' capacity to generalize their knowledge effectively and apply it to novel situations involving known entities~\cite{son2023removing}.

\begin{figure}[h]
    \centering
    \includegraphics[width=\columnwidth]{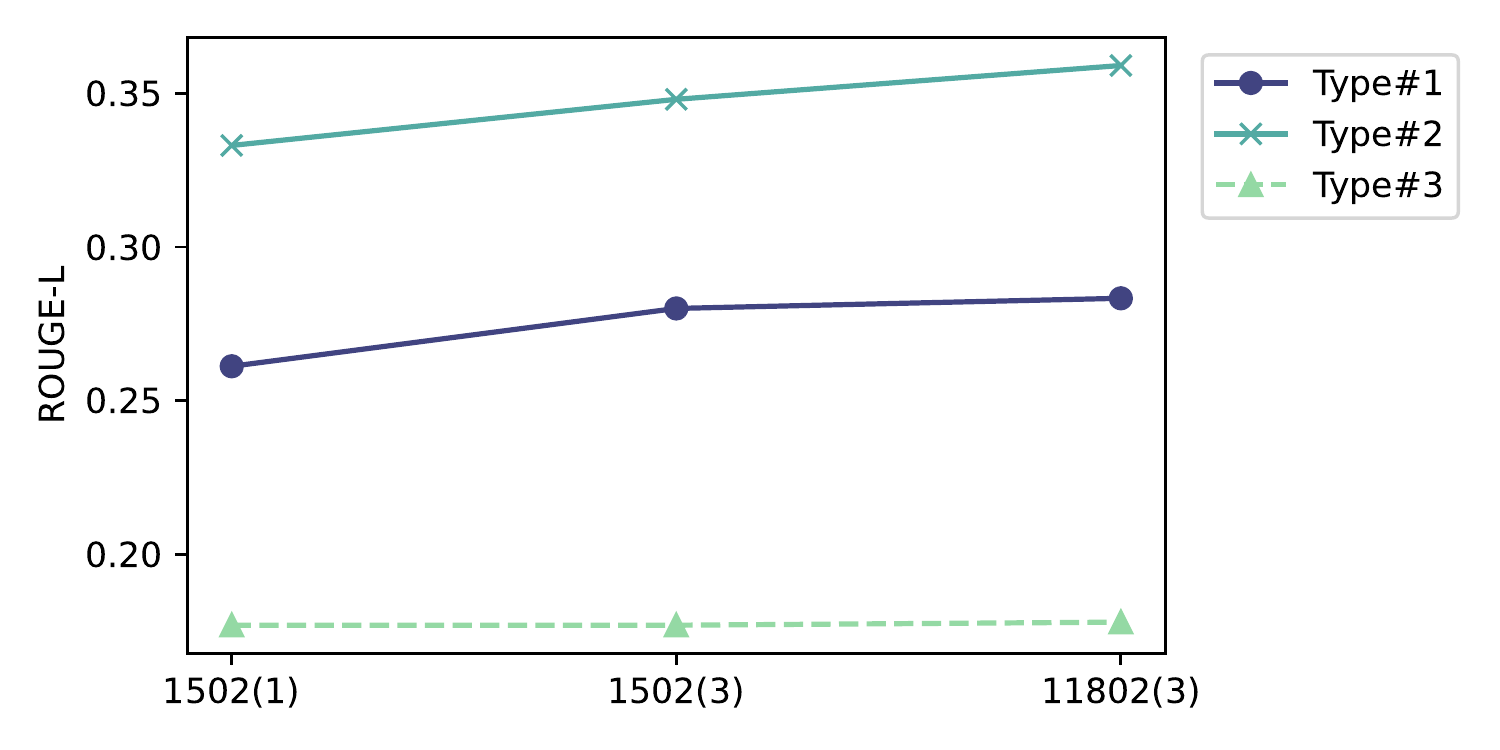}
    \caption{Performance of Vicuna across varying training steps. The x-axis denotes the training step, presented in the format sample\_size(epoch). The y-axis displays the corresponding ROUGE-L scores.}
    \label{vicuna}
\end{figure}

Furthermore, we evaluate the financial reasoning abilities of the best-performing model, instruction-tuned LLama 13B, across different dataset sizes and training steps. Specifically, we conducted experiments by training the model for (1) 3 epochs on an 11,802-sample dataset, (2) 3 epochs on a smaller 1,502-sample dataset, and (3) 1 epoch on the same 1,502-sample dataset, where each company in the full dataset was represented by 2 samples. Our results reveal that LLama's performance improved with an increasing number of training steps. However, even the model trained on the smallest configuration exhibited superior performance compared to the instruction-tuned GPT-J, which was the second-best model trained on the full dataset. These findings suggest that model size may be a critical factor in generating coherent financial reasoning, while dataset size may not be as significant.

\subsection{Human Preference}

\begin{figure}[ht]
    \centering
    \includegraphics[width=\columnwidth]{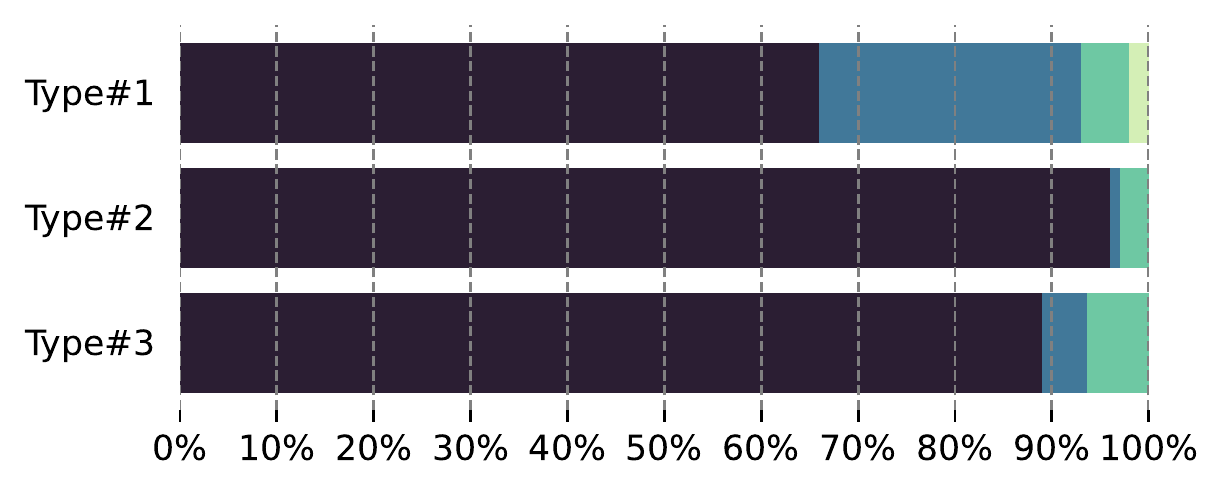}
    \caption{Human preference on generated samples. Dark Blue for LLama, Green for Galactica, Blue for GPT-J, and Yellow for Pytha(2.8B)}
    \label{human}
\end{figure}

To comprehensively evaluate the performance of each instruction-tuned model, a human preference test was conducted on their generated outputs. A panel of human evaluators was presented with four texts, each from one of the models, namely LLama, Galactica, GPT-J, and Pythia(2.8B), and asked to indicate their preference based on several factors, including coherence, relevance, and fluency. The results of the human preference test, depicted in Figure~\ref{human}, reveal that the LLama model was the most preferred choice, followed by the GPT-J model. This outcome is consistent with the findings of our previous investigation, which utilized automated metrics.

\section{Limitations and Future Work}
It is worth noting that due to hardware constraints, we were unable to investigate the emergent characteristic of financial reasoning ability on models beyond 13B parameters.  Additionally, we do not open-source expert-written samples due to copyright issues. Nevertheless, this work still represents the most comprehensive investigation to date on the behavior of language models for financial reasoning generation and the first to make a dataset for financial reasoning publicly available. Going forward, we encourage the financial natural language processing community for collaborative efforts to create larger datasets for financial reasoning tasks and to experiment with larger language models. We believe that such efforts will enable more comprehensive evaluations of language models and their potential for financial reasoning generation, ultimately advancing the state of the art in this field.

\section{Conclusion}

To the best of our knowledge, this work represents the first public effort to investigate the financial reasoning ability of language models.  Our research seeks to contribute to the understanding of the efficacy of language models in the field of finance, with a particular emphasis on their ability to engage in sophisticated reasoning and analysis within the context of investment decision-making. We confirm that the ability to generate coherent investment opinions first emerges in models with 6B parameters and scales as the model gets larger until 13B parameters. Additionally, this study introduced a novel prompting method, In-Context Question-Answering, truth-faithful generation of LLMs. The research also identified the limitations of LLMs in aligning with human evaluators for evaluating financial texts. Finally, we make a valuable contribution to the field by open-sourcing sFIOG, a dataset consisting of 11,802 synthetic investment thesis samples. 

\section{Acknowledgments}
This work was supported by the Ministry of Employment and Labor and HRD Korea's  K-Digital Platform project. 

\bibliographystyle{named}
\bibliography{ijcai23}

\appendix

\section{Sample Prompt, Q-A Pair, and Generation Example}
\label{sec:appendix}

We present a sample prompt, corresponding Q-A pair, and a generation example from our experiments, focusing on Nvidia Corporation. 

\subsection{Prompt}
\label{sec:appendix:prompt}
In our experiment, we use the following template for generations.

\begin{lstlisting}
prompt = f"Assume you are a professional financial analyst. Read the provided question and answer pair about {company} and write an investment thesis be logical and argumentative. \n QA: {QApair} Please write in English language. \n ### Investment Thesis:"
\end{lstlisting}

\subsection{Sample Question and Answer Pair}
Here we present an example of a question-and-answer pair. For better visibility, we display ten questions out of the fifteen used in our experiments.

\begin{description}
\item[Q1:] What are the primary business segments of Nvidia Corporation?
\item[A1:] Nvidia operates through two segments: GPU and Tegra Processor.
\item[Q2:] How does the Tegra brand contribute to Nvidia's portfolio?
\item[A2:] The Tegra brand integrates an entire computer onto a single chip, and incorporates GPUs and multi-core CPUs to drive supercomputing for mobile gaming and entertainment devices, as well as autonomous robots, drones, and cars.
\item[Q3:] What markets does Nvidia address with its processor platforms?
\item[A3:] Nvidia's processor platforms address four markets: Gaming, Professional Visualization, Datacenter, and Automotive.
\item[Q4:] What is the predicted earnings improvement for Nvidia in 2021 and 2022?
\item[A4:] Nvidia earnings are expected to improve to \$9.50 a share in 2021, and anticipate a +15\% increase into 2022.
\item[Q5:] What trends are causing increased demand for Nvidia's products?
\item[A5:] Nvidia is at the confluence of a couple of macro trends in the chip space, including the growth of need for data and cloud computing and the widespread surge in gaming, both accelerated by COVID-19.
\item[Q6:] How is Nvidia's acquisition of Mellanox contributing to its growth story?
\item[A6:] The Mellanox acquisition contributes to Nvidia's long term growth, despite having a negative impact on near-term margin numbers.
\item[Q7:] What is the predicted gross margin for Nvidia for FY 21 and FY 22?
\item[A7:] InSight is forecasting a gross margin of over 65\% for FY 21 and 66\% for FY 22 for Nvidia.
\item[Q8:] How do Nvidia's partnerships in machine learning and self-driving contribute to its growth prospects?
\item[A8:] NVDA’s partnerships in machine learning and self-driving could become consumer worthy in the coming decade and provide an additional growth driver not currently factored into the price target.
\item[Q9:] What is Nvidia's current stance on dividends?
\item[A9:] The dividend is currently at \$0.16 and raising it is not a priority.
\item[Q10:] How is Nvidia involved in the Medical Devices sector?
\item[A10:] Nvidia chips are being increasingly used in the healthcare diagnostic and equipment groups, adding to the company's growth.
\end{description}

\subsection{Generation Example}
\label{sec:appendix:gen}
The following section presents a sample output from the instruction-tuned LLama 13B model, our top performer in the evaluations conducted. The text was generated with a single decoding option: a maximum token limit of 512. Below, we demonstrate the model's generated investment thesis for Nvidia Corporation:

\begin{lstlisting}
Nvidia is a leading player in the semiconductor and computer technology industry, with a strong presence in the gaming, professional visualization, datacenter, and automotive markets. The company's primary business segments are GPU and Tegra Processor, which are designed to address the growing demand for data and cloud computing, as well as the increasing popularity of gaming and entertainment devices.

The acquisition of Mellanox is expected to contribute to Nvidia's long-term growth, despite having a negative impact on near-term margin numbers. The company's partnerships in machine learning and self-driving also hold significant growth potential, as these technologies are expected to become consumer worthy in the coming decade.

Nvidia is committed to producing energy-efficient chipsets and has set a goal to source 65% of its global power usage from renewable sources by 2025, making a positive contribution to the fight against climate change. The company's corporate governance is considered disciplined and transparent, with a diverse board of directors and few flags in accounting or reporting practices.

Overall, Nvidia is well-positioned to benefit from the growth in demand for data and cloud computing, as well as the increasing popularity of gaming and entertainment devices. The company's acquisition of Mellanox and partnerships in machine learning and self-driving are expected to drive long-term growth, and the company's commitment to sustainability is a positive factor. Given these factors, we believe that Nvidia is a strong investment opportunity, with a potential for a high ROI.
\end{lstlisting}

\end{document}